\documentclass[a4paper,conference]{IEEEtran}
\ifCLASSINFOpdf
\else
\fi
\usepackage{graphicx}
\graphicspath{{./images/}}

\usepackage{multirow}
\usepackage[normalem]{ulem}
\useunder{\uline}{\ul}{}

\usepackage{float}

\usepackage[english]{babel}
\usepackage{hyperref}
\addto\extrasenglish{%
}

\usepackage{booktabs}
\usepackage[para]{threeparttable}

\newcommand{\ra}[1]{\renewcommand{\arraystretch}{#1}}
\usepackage{tabularx}

\usepackage{amsmath}
\usepackage{siunitx}
\usepackage{geometry}
\usepackage{array}
\lccode`8=`8
\begin{document}
%
% paper title
% Titles are generally capitalized except for words such as a, an, and, as,
% at, but, by, for, in, nor, of, on, or, the, to and up, which are usually
% not capitalized unless they are the first or last word of the title.
% Linebreaks \\ can be used within to get better formatting as desired.
% Do not put math or special symbols in the title.

\title{ESResNet: Environmental Sound Classification Based on Visual Domain Models}

% author names and affiliations
% use a multiple column layout for up to three different
% affiliations
\author{
\IEEEauthorblockN{
Andrey Guzhov\textsuperscript{1,2},
Federico Raue\textsuperscript{2},
J{\"o}rn Hees\textsuperscript{2},
Andreas Dengel\textsuperscript{1,2}
}
\IEEEauthorblockA{
\textsuperscript{1}DFKI GmbH\\
\textsuperscript{2}TU Kaiserslautern \\
Kaiserslautern, Germany \\
\url{firstname.lastname@dfki.de}
}
}
% conference papers do not typically use \thanks and this command
% is locked out in conference mode. If really needed, such as for
% the acknowledgment of grants, issue a \IEEEoverridecommandlockouts
% after \documentclass

% for over three affiliations, or if they all won't fit within the width
% of the page, use this alternative format:
%
%\author{\IEEEauthorblockN{Michael Shell\IEEEauthorrefmark{1},
%Homer Simpson\IEEEauthorrefmark{2},
%James Kirk\IEEEauthorrefmark{3},
%Montgomery Scott\IEEEauthorrefmark{3} and
%Eldon Tyrell\IEEEauthorrefmark{4}}
%\IEEEauthorblockA{\IEEEauthorrefmark{1}School of Electrical and Computer Engineering\\
%Georgia Institute of Technology,
%Atlanta, Georgia 30332--0250\\ Email: see http://www.michaelshell.org/contact.html}
%\IEEEauthorblockA{\IEEEauthorrefmark{2}Twentieth Century Fox, Springfield, USA\\
%Email: homer@thesimpsons.com}
%\IEEEauthorblockA{\IEEEauthorrefmark{3}Starfleet Academy, San Francisco, California 96678-2391\\
%Telephone: (800) 555--1212, Fax: (888) 555--1212}
%\IEEEauthorblockA{\IEEEauthorrefmark{4}Tyrell Inc., 123 Replicant Street, Los Angeles, California 90210--4321}}

% use for special paper notices
%\IEEEspecialpapernotice{(Invited Paper)}

% make the title area
\maketitle

% As a general rule, do not put math, special symbols or citations
% in the abstract
% But I want to put it :p
\begin{abstract}

Environmental Sound Classification (ESC) is an active research area in the audio domain and has seen a lot of progress in the past years.
However, many of the existing approaches achieve high accuracy by relying on domain-specific features and architectures, making it harder to benefit from advances in other fields (e.g., the image domain).
Additionally, some of the past successes have been attributed to a discrepancy of how results are evaluated (i.e., on unofficial splits of the UrbanSound8K (US8K) dataset), distorting the overall progression of the field.

The contribution of this paper is twofold.
First, we present a model that is inherently compatible with mono and stereo sound inputs.
Our model is based on simple log-power Short-Time Fourier Transform (STFT) spectrograms and combines them with several well-known approaches from the image domain (i.e., ResNet, Siamese-like networks and attention).
We investigate the influence of cross-domain pre-training, architectural changes, and evaluate our model on standard datasets.
We find that our model out-performs all previously known approaches in a fair comparison by achieving accuracies of 97.0\,\% (ESC-10), 91.5\,\% (ESC-50) and 84.2\,\% / 85.4\,\% (US8K mono / stereo).

Second, we provide a comprehensive overview of the actual state of the field, by differentiating several previously reported results on the US8K dataset between official or unofficial splits.
For better reproducibility, our code (including any re-implementations) is made available.

\end{abstract}

% no keywords

% For peer review papers, you can put extra information on the cover
% page as needed:
% \ifCLASSOPTIONpeerreview
% \begin{center} \bfseries EDICS Category: 3-BBND \end{center}
% \fi
%
% For peerreview papers, this IEEEtran command inserts a page break and
% creates the second title. It will be ignored for other modes.
\IEEEpeerreviewmaketitle

\section{Introduction} \label{sec:intro}
With the increasing popularity of voice assistants, many of which use Deep Learning techniques, the currently most apparent task from the audio domain is probably automatic speech recognition.
However, apart from this very prominent example, many other challenges still exist in the audio domain.
One of these challenges is Environmental Sound Classification (ESC), which is concerned with correctly differentiating between sound classes that we experience in our everyday environment (e.g., ``baby crying'', ``car honking'', ``children playing'', ``dog barking'', ``siren'', ``snoring'', ``street music'').
While ESC has many potential application areas, one of the most obvious ones is multimedia retrieval, in which ESC could be used to improve the performance of video retrieval systems by making better use of the audio modality \cite{jin2016video}.
Another application area is the automated analysis of urban sounds, for example to offer more detailed insights for high noise levels \cite{raimbault2005urban}.

While ESC is a comparably young field, a lot of progresses were made after great datasets such as the ESC-50~\cite{piczak2015esc} and \mbox{UrbanSound8K} (US8K)~\cite{salamon2014us8k} found wide acceptance in the community.
However, we observed that the general trend in the ESC community is to design audio-domain-specific architectures and combine them with specially engineered features.
On one side, this approach makes more difficult to benefit from advances made in other fields, such as the computer vision community.
On the other, this scenario sparked our interest to investigate how well a current \mbox{state-of-the-art} approach from the image domain would perform on ESC.
During our investigations, we found, that while our approach immediately out-performed all previous ones on the ESC-50 dataset, it initially seemed to perform quite poorly on US8K, despite the fact that it can actually make use of stereo inputs.
However, during our follow-up, we noticed that there are reproducibility problems wrt. prior publications reporting on the US8K dataset: many existing approaches lack necessary details for reproduction and about the used dataset splits. 
Given a fair comparison, our approach in fact out-performs all prior ESC models also on the US8K.

The remainder of this paper is organized as follows.
In \autoref{sec:related} we discuss prior models that were used for environmental sound classification.
We then describe our proposed approach based on log-power STFT spectrograms and a well-known CNN model in \autoref{sec:model}, how it was trained and evaluate in \autoref{sec:exp}, before presenting our results in \autoref{sec:results} and concluding with a summary and future work in \autoref{sec:conc}.

\section{Related Work} \label{sec:related}

Unlike image-related tasks (image classification, segmentation, object detection, etc.), the environmental sound classification task implies the usage of locally correlated one-dimensional signals, so the input is stretched along a single axis.
The most widely known datasets in the field of environmental sound classification are the \mbox{ESC-50\:/\:-10} \cite{piczak2015esc} and the UrbanSound8K (US8K)~\cite{salamon2014us8k}, further detailed in \autoref{sec:exp}.

The representation of audio is quite different from visual signals (e.g., photo) that have local correlations in both spatial dimensions.
Thus, many methods were proposed specifically tailored towards the audio domain.
We can divide them into the following major groups.
A comprehensive overview and comparison of all methods can be found in the \autoref{tbl:results}.

\subsection{Raw Waveform and 1D-CNN} \label{ssec:raw_cnn}
The use of a raw signal as an input provides a straightforward solution to build a model that handles any sort of time-frequency transformation internally.
The most important property of this class of models is that data pre-processing is not needed.
First, \cite{tokozume2017envnet} and \cite{tokozume2017envnetv2} proposed a one-dimensional architecture called \mbox{EnvNet v1 / 2} that was able to achieve state-of-the-art results at that time.
Later, in \cite{zhu2018multires} the concept of 1D-CNNs was extended into a model that operated on an input signal at different time scales.
Another way to improve performance of this type of models was proposed in \cite{abdoli2019cnn1d} where the use of gammatone filterbanks for the initialization of model allowed to improve results in comparison to the otherwise random weight initialization.

In contrast, for simplicity and to reduce the amount of trainable parameters (considering limited training data), we decided to rely on a fixed time-frequency transform with a wide spectrum range in our model.
However, we mention potential improvements in this direction in our future work.

\subsection{Learnable Filterbanks and 2D-CNN} \label{ssec:learn_fb}
While one-dimensional CNNs handle all transformations of input signal internally, which makes it possible to apply them in an end-to-end fashion, this approach involves a lack of control over the transformed representation.
The use of gammatone initialization in \cite{abdoli2019cnn1d} helped to overcome this issue partially.
The uniqueness of \cite{sailor2017convrbm} is that the model was split into two parts, namely Convolutional Restricted Boltzman Machines (ConvRBM) for feature extraction (instead of the fixed procedure that is used in our work) and the CNN proposed in \cite{piczak2015cnn} for the actual classification task.

\subsection{Pre-computed Time-Frequency Representation and 2D-CNN} \label{ssec:tf_cnn}
The first model that set the baseline for the environmental sound classification was the CNN proposed in \cite{piczak2015cnn} (Piczak-CNN) that operated on Mel-scaled \cite{volkmann1937mel} spectrograms.
The use of a fixed feature extraction procedure made it possible to obtain a model's input that possessed the required characteristics.
Further development of single-feature input and research on data augmentation techniques were done in \cite{salamon2017cnn}.
Follow-up studies involved the extension of input features to others that based mostly on the Short-Time Fourier Transform (STFT) \cite{allen1977stft} to Mel-Frequency Cepstral Coefficients (MFCC) \cite{logan2000mel}, \cite{boddapati2017classifying}; Cross Recurrence Plot (CRP) \cite{marwan2002crp}, \cite{boddapati2017classifying}; Teager's Energy Operator (TEO) \cite{kaiser1993teo}, \cite{agrawal2017teo}; (Phase-Encoded) FilterBank Energies ((PE)FBE) \cite{tak2017pefbe}; gammatone-spectrogram \cite{slaney1993gammatone}, \cite{zhang2018mixup}, \cite{zhang2019crnn}; chromagram \cite{shepard1964chroma}, \cite{su2019tscnnds}; spectral contrast \cite{jiang2002spectral}, \cite{su2019tscnnds}; and Tonnetz \cite{harte2006tonnetz}, \cite{su2019tscnnds}.

However, all of the aforementioned features were developed with a reduction of computational complexity or compression in mind.
With the growth of computational capacity, it seems that we can now make use of a single-feature that covers the full range without any reduction.

The model we propose belongs to this major group, it is also handling single-feature input (log-power spectrograms).

We provide detailed description of features used in the aforementioned studies in the \autoref{tbl:results}.

\subsection{Data Augmentation} \label{sec:related:aug}
Data augmentation is a powerful technique that allows to increase variability in the training data and thus acts as a regularizer preventing overfitting.
According to \cite{salamon2017cnn} and \cite{tokozume2017envnetv2} there are the following transformations that augment audio training data:
\subsubsection{Time Stretching}
This method changes the duration of the audio, while keeping its spectral characteristic untouched.
\subsubsection{Pitch Shift}
In opposite to time stretching, this method allows to manipulate spectral characteristics and preserve duration of the track.
\subsubsection{Time Inversion}
Time inversion that was applied in \cite{tokozume2017envnetv2} is an effective data augmentation technique that is related to random flip of images during the training on the visual classification datasets.

\subsection{Comparability of Results on the UrbanSound8K (US8K) Dataset}
According to our findings, there are at least five papers (three in the 2019), whose reported results are not directly comparable with others.
In particular, as reported by \cite{zhang2018mixup}, the authors of \cite{agrawal2017teo} used an unofficial split of the US8K dataset.
Also, the authors of \cite{boddapati2017classifying} stated that the results were obtained on a non-standard split, whereas the authors of \cite{abdoli2019cnn1d} provide the description of a custom snippets generation strategy.
Finally, we determined that results published by \cite{su2019tscnnds} and \cite{wang2019tfnet} are incomparable with those acquired on the official split \cite{salamon2014us8k} of the US8K dataset as well.
We provide further details on this in the \autoref{tbl:results} and \autoref{ssec:off_vs_unoff}.

\section{Model} \label{sec:model}

In this paper, we propose a visual domain convolutional neural network in conjunction with log-power spectrograms to solve the environmental sound classification task.
This section describes the architecture of the model and how it is extended by the attention mechanism.
We also describe its application to stereo audio, the initialization of the network's weights and the process of log-power spectrogram computations.

\subsection{Residual Networks} \label{ssec:resnet}
\begin{figure}[hbt]
\centering
\includegraphics[trim={0.75cm 1cm 12.5cm 10cm},clip,width=\linewidth]{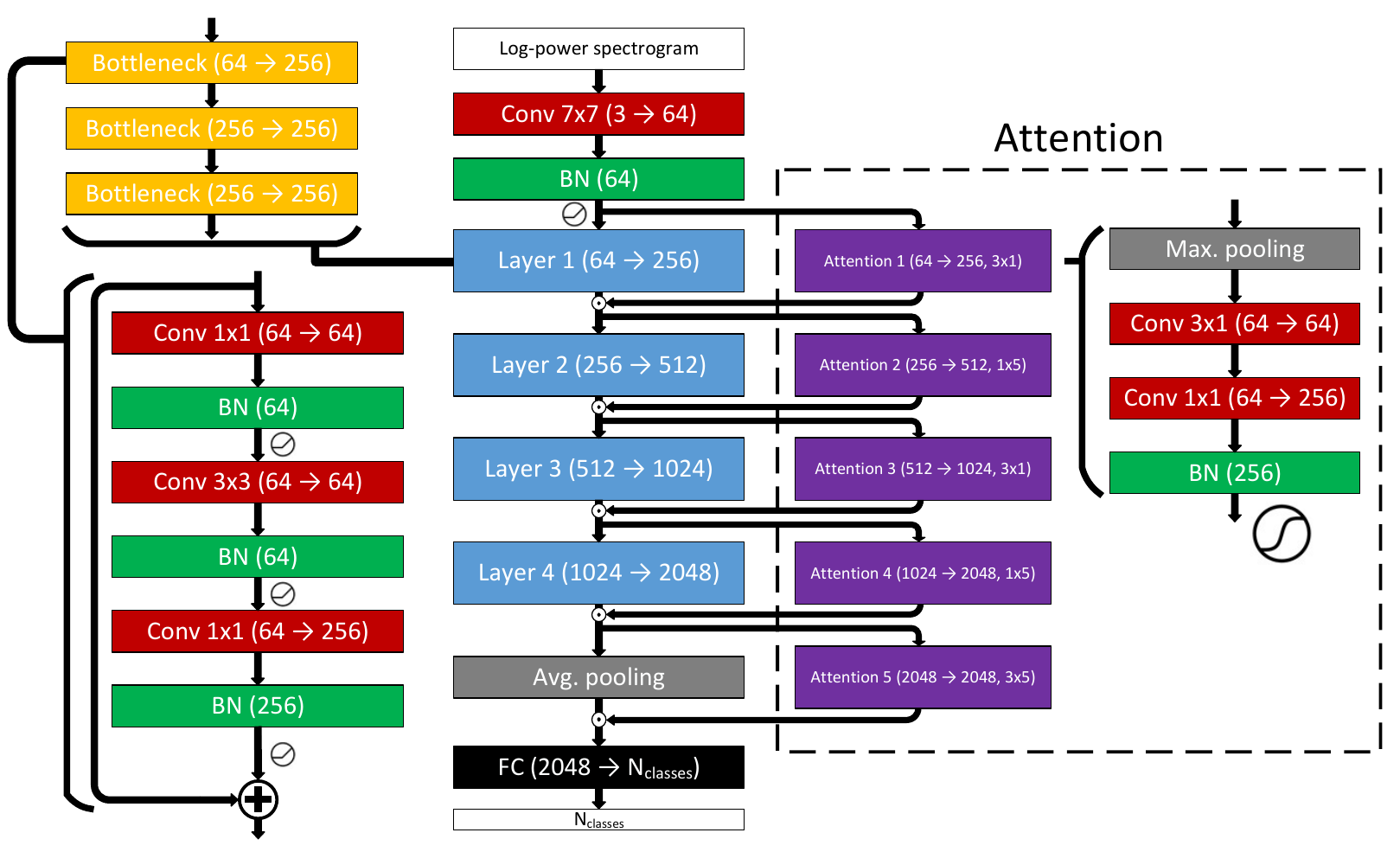}
\caption{Overview of the ESResNet model handling single-channel input. The main branch (2nd column) consists of the Convolutional layer (red) stacked together with the Batch Normalization layer (green), followed by the residual layers 1\,--\,4 (blue), the Average pooling (gray), and the Fully-Connected layer (black). On the left, the typical structure of a residual layer is presented. Each residual layer consists of the stack of the bottleneck layers (orange) that include \mbox{Conv-BN}-operations applied sequentially and the skip-connection. Rectified Linear Unit (ReLU) serves as an activation function. On the right (bounded by the dashed line), the optional extension of the \mbox{ESResNet} model by the attention blocks is presented. If applied, the attention block (violet) is stacked in parallel to the residual layer 1 to 4 or to the average pooling layer. The attention block includes the Max-pooling operation (gray) followed by the depth-wise separable convolution stacked together with batch normalization. The output of the attention block is given by the logistic function.}
\label{fig:esresnet_a}
\end{figure}

Residual neural networks are characterized by the additional skip connections that bypass some of the layers and merge their input and output.
The motivation for this is to prevent gradient vanishing that made it very difficult to design deep neural networks before \cite{he2016resnet}.
In our work, we propose the \mbox{ESResNet} model based on the vanilla \mbox{ResNet-50} architecture in order to demonstrate its ability to achieve \mbox{state-of-the-art} results on a domain the model was not designed for.
The overall structure of the model is presented in \autoref{fig:esresnet_a}.

\subsection{Attention} \label{ssec:attention}
The attention mechanism was presented initially for the use in conjunction with recurrent models, in particular, in sequence modelling tasks \cite{vaswani2017attention}.
The main goal of it was to highlight relevant parts of a long sequence and to get rid of irrelevant ones.
In the visual domain, one uses attention blocks in order to produce weighting for the input signal.
Usually, there are several attention sub-branches consisting of one or many convolutional layers that process feature maps in parallel with the main branch.

For the environmental sound classification task, the main purpose of the attention blocks is to focus the model on the most important information in both the time and frequency domain.
To implement the attention mechanism, we extended our \mbox{ESResNet} model (inspired by \cite{zhang2019crnn}), by adding a stack of attention blocks in parallel (\mbox{ESResNet-Attention}, \autoref{fig:esresnet_a}).
Each block among the first 4 handles either frequency- or time-related information.
For instance, the first attention block $A_{1}$ receives the same input as the first layer $L_{1}$, then it processes the signal $x$ using frequency-dedicated convolutional filters and provides an output of the same shape as the one provided by the $L_{1}$.
Finally, the input of the second layer is constructed by the element-wise multiplication of outputs of $L_{1}$ and $A_{1}$ blocks (\autoref{eq:att}).

\begin{equation}
    L_{\emph{i}}^{\emph{att.}}(x) = L_{i}(x) \odot A_{i}(x)
    \label{eq:att}
\end{equation}

The last attention block handles a joint time-frequency representation.
The core of the attention block is a depth-wise separable convolution \cite{chollet2017xception}.
Output of each attention block is given by the logistic function.

\subsection{Spectrogram} \label{ssec:spec}
A spectrogram is an image-like representation of the spectrum of frequencies varying with time.
In relation to digital signal processing, there are several ways to obtain a spectrogram.
It can be generated using filterbanks, Fourier (or more generally wavelet) transform, etc.
In our work, we compute log-power spectrogram $S$ from the STFT of an audio signal $X(\tau, \omega)$ (\autoref{eq:spec}).

\begin{equation}
    S=10\:Log_{10}|X(\tau, \omega)|^2
    \label{eq:spec}
\end{equation}

\subsubsection{Short-Time Fourier Transform (STFT)}
STFT belongs to the family of Fourier-related transforms and is used to determine magnitude and phase of basis sinusoidal frequencies $\omega$ at different time points $\tau$ in a time-domain signal $x$.

\begin{equation}
    X(\tau, \omega) = \sum_{n = -\infty}^{\infty} x[n]w[n - \tau] e^{-j \omega n}
    \label{eq:stft}
\end{equation}

In practice, to compute \autoref{eq:stft}, one splits input signal into overlapping frames multiplied by window function $w$, then the Fast Fourier Transform (FFT) \cite{cooley1965fft} is being applied to each frame separately.

\subsubsection{Window Function}
In order to reduce spectrum perturbances caused by the framing, a window function is applied.
The use of windowing reduces the amount of noise in the spectrum and therefore improves the signal-to-noise ratio.
The drawback of the usage of a window function is so-called spectral leakage.
Spectral leakage is a common name for the non-zero values produced by Fourier transform at frequencies other than fundamental.
The choice of window function is a trade-off between many characteristics.
In our work, we decided to choose the minimum \mbox{4-term} \mbox{Blackman-Harris} window \cite{harris1978window} which is given by \autoref{eq:bh92} as it provides reasonable bandwidth and very low spectral leakage making it a good choice as a \mbox{general-purpose} window \cite{heinzel2002spectrum}:

\begin{equation}
\begin{split}
    w{[}k{]} = &\   a_0 - a_1\,cos\bigg(\frac{2 \pi k}{N}\bigg) +  a_2\,cos\bigg(\frac{4 \pi k}{N}\bigg) \\ & - a_3\,cos\bigg(\frac{6 \pi k}{N}\bigg)
\end{split}
\label{eq:bh92}
\end{equation}

\begin{center}
    $a_0=0.35875;a_1=0.48829;a_2=0.14128;a_3=0.01168$
\end{center}

As a trade-off between time and frequency resolution, we split the input signal into frames of $37.5\:\si{\milli\second}$ length.
The corresponding overlap between subsequent frames depends on the chosen window function.
In our case, the recommended overlap for the \mbox{Blackman-Harris} window of $66.1\,\%$ ($24.8\:\si{\milli\second}$) \cite{heinzel2002spectrum} was used.

\subsection{Input Channel Transformation} \label{ssec:align}

For image classification models, such as ours, the usual way to represent input data is an RGB model with 3 input channels (red, green and blue).
However, our spectrograms only provide input in form of a single-channel (grayscale values).
One way to tackle this issue is to replicate the spectrogram to other channels or to pass zeros instead.
The major drawback of this solution is either unnecessary redundancy or the loss of information, and increased computational cost.

\begin{figure}[hbt]
\centering
\includegraphics[trim={0.5cm 10.5cm 11.5cm 0.5cm},clip,width=\linewidth]{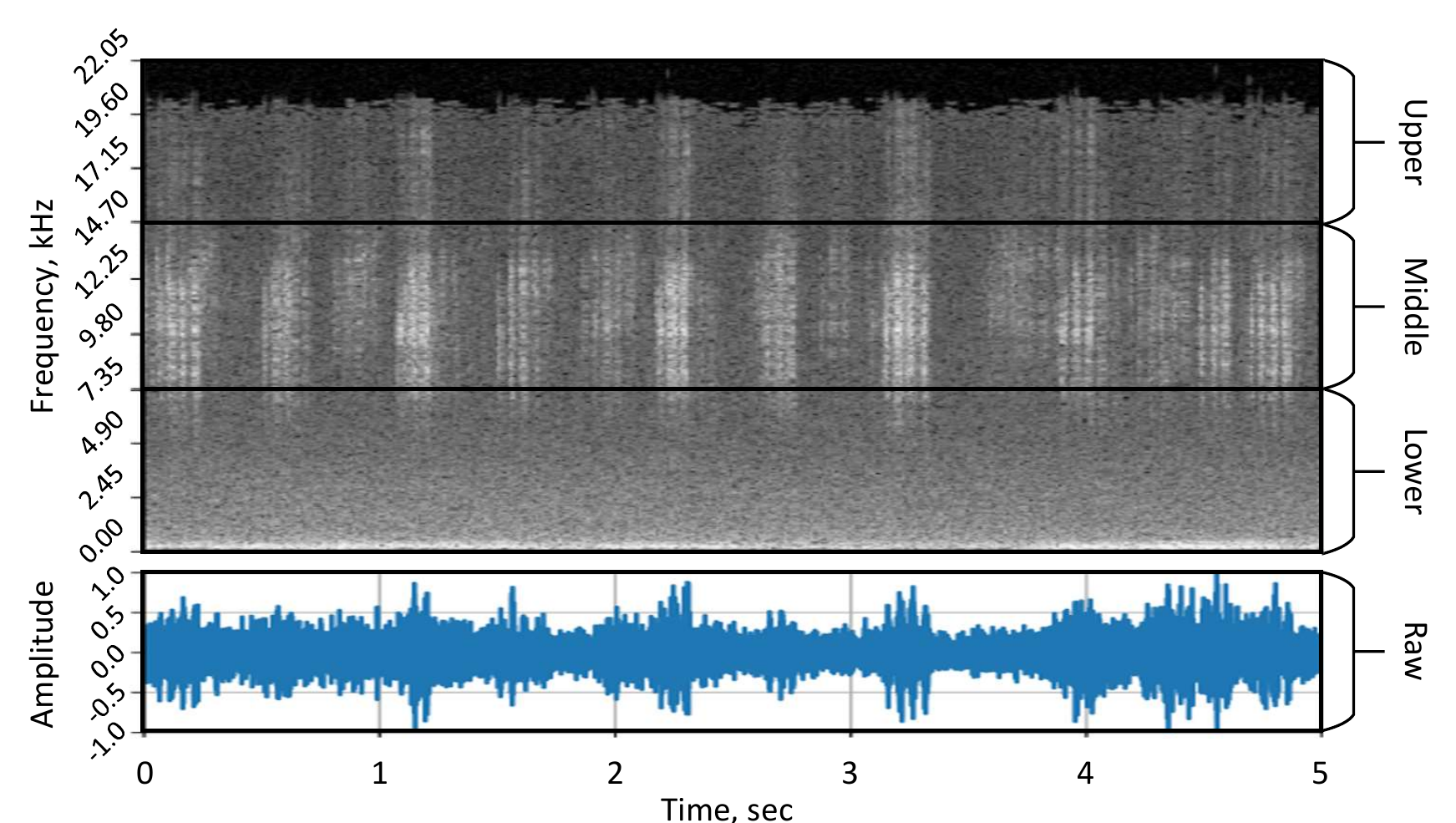}
\caption{Example of the ESResNet model's input. The raw audio (bottom) is used to create the log-power spectrogram that is split into three frequency bands: lower, middle and upper. Thus, the network's input consists of three channels that are aligned with visual channels (red, green and blue, respectively).}
\label{fig:spec}
\end{figure}

In order to overcome this limitation, we decided to map the spectrogram along its frequency axis onto the three input channels, so it is split into 3 frequency bands (\autoref{fig:spec}): lower ($0.00-7.35\:\si{\kilo\hertz}$), middle ($7.35-14.70\:\si{\kilo\hertz}$) and upper ($14.7-22.05\:\si{\kilo\hertz}$).

\subsection{Handling Stereo Audio Using Siamese-like Architecture} \label{ssec:siamese}

The way how humans perceive audible information is inherently stereo.
In this work, we exploit advantages brought by additional audio channels and show on the US8K dataset that a minor architectural tweak helps us to out-perform \mbox{state-of-the-art} results.

\begin{figure}[hbt]
\centering
\includegraphics[trim={0.5cm 12cm 14cm 1cm},clip,width=\linewidth]{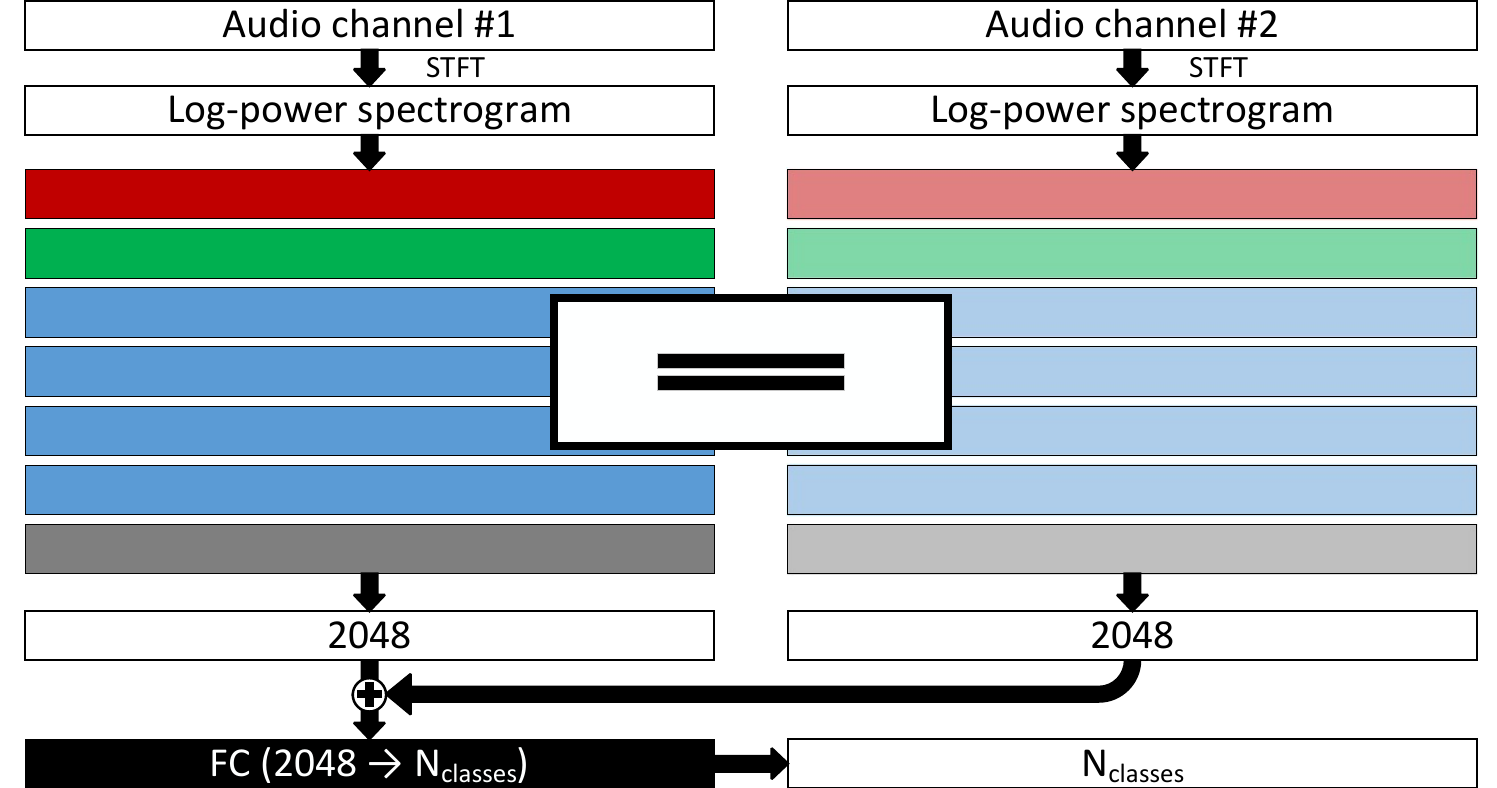}
\caption{Overview of the ESResNet model handling two-channel input.
Optional attention blocks are omitted.
The color scheme is the same as in \autoref{fig:esresnet_a}.
The first channel is passed through the main branch (saturated coloring), so the resulting embedding of size 2048 is acquired.
Then, the second channel is passed through the same set of weights (pale coloring), resulting in another embedding of size 2048.
Finally, the two embeddings are added element-wise and their sum is passed through the last fully-connected layer that performs the final prediction of a class the input sample belongs to.}
\label{fig:siamese}
\end{figure}

Siamese neural networks were developed to produce a similarity measure for two input samples \cite{koch2015siamese}.
In this work, we however use the common broader notation to call any network Siamese that applies the same set of weights to two different inputs and thus produces two comparable vectors (or embeddings).
As \autoref{fig:siamese} illustrates, we take a two-channel audio input followed by log-power spectrogram computation (via STFT) and pass each channel separately through the layers.
After we obtain the network's outputs, we fuse them by an element-wise addition and pass the resulting embedding through the last fully-connected layer that performs the final classification.

\subsection{ImageNet Training as Weight Initializer} \label{ssec:imagenet}
The ESC datasets contain a limited number of samples.
This setup becomes especially important in case of the \mbox{ESC-50} dataset as it provides the challenging task to distinguish between 50 classes using only 1600 training samples \cite{piczak2015esc}.

To leverage the full power of deep neural networks, the amount of data should grow exponentially with the amount of parameters.
If the number of training samples is restricted, one way is to perform fine-tuning.
In this work, we decided to employ a model that was trained from scratch on the ImageNet dataset \cite{deng2009imagenet}.
The ImageNet dataset provides more than 1 million training samples divided into 1000 classes.
As we will see in \autoref{sec:results}, the initialization of weights based on a pre-training on the ImageNet image classification task is beneficial for the environmental sound classification.

\section{Experimental Setup} \label{sec:exp}
In this section we describe the setup of our experiments, starting with the datasets, their pre-processing and how our model was trained.
We also describe our reproduction of previous results / re-implementation of their approaches for comparison.

\subsection{Datasets} \label{ssec:datasets}
\subsubsection{\mbox{ESC-50 / -10}}
The \mbox{ESC-50} dataset consists of 2000 monarual samples belonging to 50 classes that can be divided into 5 groups, such as \emph{animal} sounds, \emph{natural and water} sounds, \emph{non-speech human} sounds, \emph{interior} and \emph{exterior} sounds \cite{piczak2015esc}.
Samples are distributed equally among classes, thus each category consists of 40 recordings.
Each track has length of 5 seconds, the native sample rate is $44.1\:\si{\kilo\hertz}$.
The dataset was divided into 5 folds by its authors that we used in current work to perform our evaluation.
The \mbox{ESC-10} dataset is a subset of the \mbox{ESC-50} dataset.
It consists of only 10 classes that are restricted to the following categories: \emph{transient / percussive} sounds with temporal patterns, sounds with strong \emph{harmonic} content, and \emph{noise / soundscapes}.
All other characteristics of the \mbox{ESC-10} dataset are equal to those of the \mbox{ESC-50} dataset.

\subsubsection{UrbanSound8K} \label{sssec:us8k}
The US8K dataset consists of 8732 samples (both mono and stereo) belonging to 10 classes: ``air conditioner'', ``car horn'', ``children playing'', ``dog bark'', ``drilling'', ``engine idling'', ``gun shot'', ``jackhammer'', ``siren'', and ``street music'' \cite{salamon2014us8k}.
The classes are not balanced in terms of overall recording lengths per class.
Each track has variable length up to 4 seconds, the native sample rate varies from $16\:\si{\kilo\hertz}$ to $48\:\si{\kilo\hertz}$.
The dataset was divided into 10 folds by its authors that we used in current work to perform our evaluation.

\begin{figure}[!hbt]
\centering
\includegraphics[trim={0.5cm 8.5cm 8.5cm 0.5cm},clip,width=\linewidth]{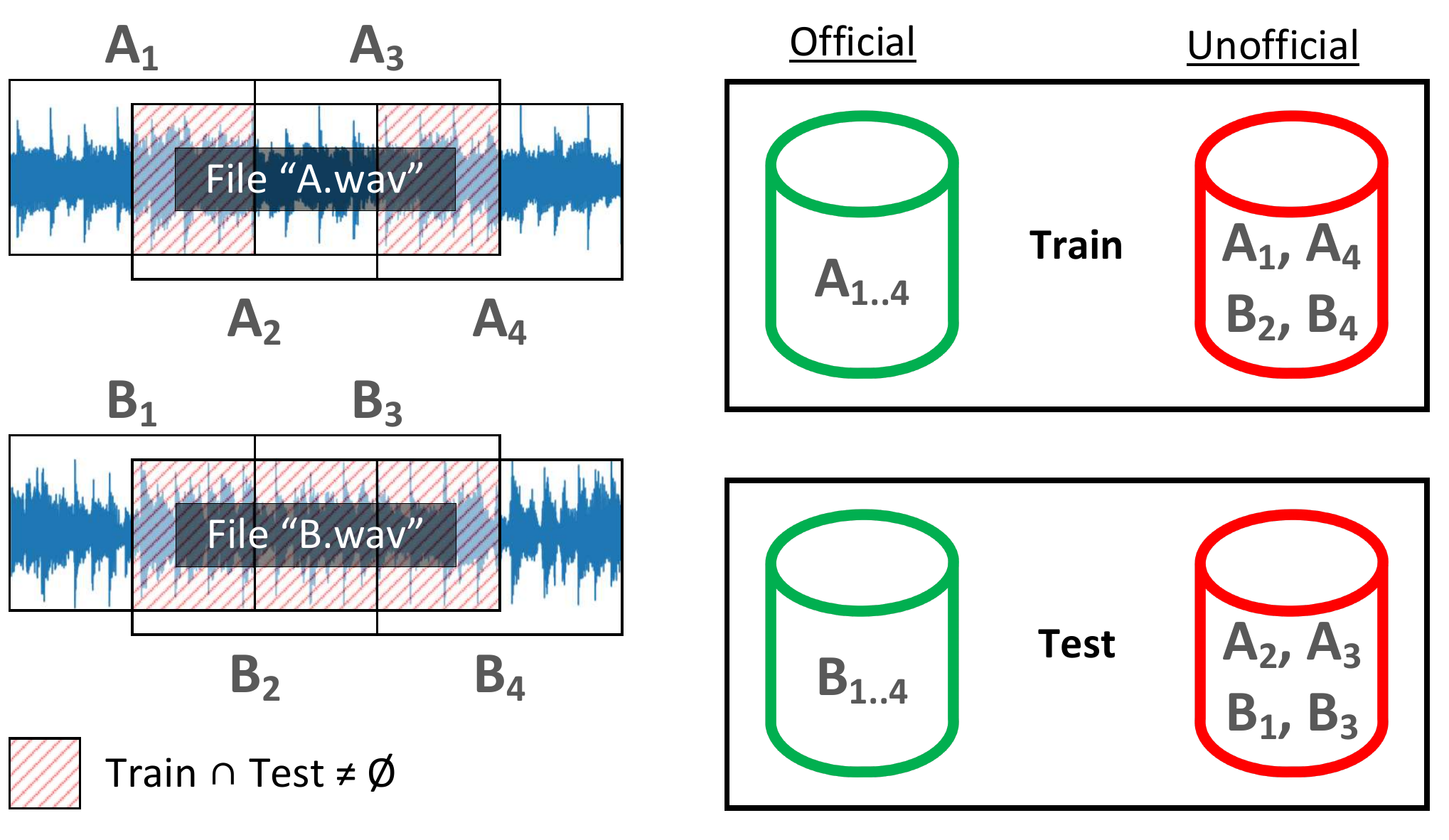}
\caption{Generation and split of the UrbanSound8K dataset. The dataset consists of the samples that were obtained applying a sliding window that had an overlap of 50\:\% to the original recordings provided by the \href{https://freesound.org/}{Freesound} project \cite{freesound2013}.
The official split of the dataset takes care of putting subsequent samples into either the train- or the test-set, so there are no shared parts of the original track in both.
A random split (frequent in unofficial splits) does not follow this constraint, which results in a mix-up of train- and test-data (highlighted area) and thus unreasonably high model performance.}
\label{fig:urbansound8k}
\end{figure}

We would like to explicitly highlight the importance of using the officially provided folds by describing the way training samples were acquired by authors of \cite{salamon2014us8k}.
As at the time of collection the number of qualitatively labeled recordings provided by the \href{https://freesound.org/}{Freesound} project \cite{freesound2013} was restricted \cite{salamon2014us8k}, each track was split into snippets that had an overlap of 50\,\% \cite{salamon2014us8k}.
Let us consider two tracks A and B belonging to the same class (\autoref{fig:urbansound8k}).
Applying a sliding window and moving it with the overlap of 50\:\%, we obtain snippets called A\textsubscript{1--4} and B\textsubscript{1--4}, respectively.
Two subsequent snippets share a part of the original track, so one has to make sure that they are presented either in training or evaluation set (the official split) and not in both (as happens by random shuffling, which is underlying many unofficial splits).

\subsubsection{Data Pre-processing}
For all datasets, audio samples were normalized to the sample rate of $44.1\:\si{\kilo\hertz}$ using Librosa 0.7.2 library \cite{librosa2020}.
According to the chosen window length of $37.5\:\si{\milli\second}$, the frame length was 1654 samples.
As the underlying implementation includes an FFT, frames were explicitly padded to the next power of 2 ($2^{11}=2048$) using a reflection padding strategy \cite{cooley1965fft}.

\subsection{Model Training} \label{ssec:model_training}

The model was trained using the Adam \cite{kingma2014adam} optimizer for 300 epochs.
The batch size was set to 16, training samples were shuffled between batches after every epoch.
During the training, the learning rate was adjusted according to an exponential decay schedule with warm-up \cite{goyal2017warmup}.
The basic learning rate value was set to $0.00025$.
For the first 5 epochs it was ten times lower, then the learning rate grew up linearly during the next 10 epochs.
After the warm-up period, it decayed exponentially with $\gamma = 0.985$ so the training ended up with learning rate of $3.37\mathrm{e}{-6}$.
In order to introduce more stability into the training process, a weight decay with $\alpha = 0.0005$ \cite{krizhevsky2012alexnet} was applied.
Other hyper-parameters \cite{kingma2014adam} such as $\beta_1$, $\beta_2$, $\epsilon$ were set to their default values.
Categorical cross-entropy served as a loss function.

During the training phase, the following augmentations were applied (see \autoref{sec:related:aug}): random time inversion and time scaling  \cite{tokozume2017envnetv2}.
The later can be considered as a combination of time stretching and pitch shift.
The main advantage of such combined transformation is its computational cheapness in comparison to aforementioned ones.
For instance, pitch shift implies forward and inverse STFT which makes it inefficient to apply this transformation on-the-fly during the training.
The probability of the time inversion was set to $0.5$.
The scaling factor was sampled uniformly from the continuous range $[1.25^{-1}, 1.25]$.

\subsection{Re-implementation} \label{ssec:reimpl}

As the results reported by \cite{wang2019tfnet} and especially \cite{su2019tscnnds} were very high, we focused on them to find the key to a such high performance.
As the description of models and\:/\:or setups did not allow us to determine the crucial component, we decided to reproduce their results.
Sadly, the authors of \cite{su2019tscnnds} did not publish their code, nor did any of them respond to our email within a month.
Hence, we re-implemented the part of the \mbox{TSCNN-DS} model called \mbox{LMCNet} and evaluated it on the official \cite{salamon2014us8k} and unofficial random split of the US8K dataset using all available implementation details provided by authors.

The authors of \cite{wang2019tfnet} had published parts of their source code including hyper-parameters for their \mbox{TFNet} model (only ESC-50), which allowed us to reproduce their results with minor re-implementations (by extending to US8K).
Sadly, after contacting them by email, the original repository disappeared.

For our evaluations on an unofficial random split of the US8K dataset we used a \texttt{StratifiedKFold} as provided by \href{https://scikit-learn.org}{scikit-learn} \cite{pedregosa2011sklearn}.
The number of splits was set to 10, all experiments were conducted with the same random seed.

We report the reproduced results in \autoref{tbl:results} (emphasized by \emph{italic} font) and discuss them in \autoref{ssec:off_vs_unoff}.

\section{Results} \label{sec:results}

\begin{table*}[hbt]
\begin{threeparttable}[t]
\caption{Evaluation Results (Accuracy, \%)}
\label{tbl:results}
\ra{1.2}
\begin{tabularx}{\textwidth}{@{}cXclcccc@{}}
\toprule
 & \multicolumn{1}{l}{\multirow{2}{*}{Model}} & \multicolumn{1}{c}{\multirow{2}{*}{Source}} & \multicolumn{1}{l}{\multirow{2}{*}{Representation}} & \multicolumn{1}{l}{\multirow{2}{*}{ESC-10}} & \multicolumn{1}{l}{\multirow{2}{*}{ESC-50}} & \multicolumn{1}{c}{US8K} & \multicolumn{1}{c}{US8K} \\
 & \multicolumn{5}{c}{} & \multicolumn{1}{l}{official} & \multicolumn{1}{l}{unofficial} \\
\midrule
\parbox[t]{2mm}{\multirow{28}{*}{\rotatebox[origin=c]{90}{\underline{Others}}}} & Human (2015) & \cite{piczak2015esc} & -- & 95.70 & 81.30 & -- & --\;\; \\
\cmidrule(l){2-8} 
 & \multicolumn{1}{l}{\textbf{Raw waveform and 1D-CNN}} & \multicolumn{6}{l}{} \\
 & EnvNet (2017) & \cite{tokozume2017envnet} & raw & 88.10 & 74.10 & 71.10 & --\;\; \\
 & EnvNet v2 (2017)  & \cite{tokozume2017envnetv2} & raw & 91.30 & 84.70 & 78.30 & --\;\; \\
 & Multiresolution 1D-CNN (2018)  & \cite{zhu2018multires} & raw & -- & 75.10 & -- & --\;\; \\
 & Gammatone 1D-CNN (2019)  & \cite{abdoli2019cnn1d} & raw & -- & -- & -- & 89.00\,\tnote{1} \\
\cmidrule(l){2-8} 
 & \multicolumn{1}{l}{\textbf{Learnable filterbank and 2D-CNN}} & \multicolumn{6}{l}{} \\
 & Piczak-CNN + ConvRBM (2017) & \cite{sailor2017convrbm} & FBE & -- & 86.50 & -- & --\;\; \\
\cmidrule(l){2-8} 
 & \multicolumn{1}{l}{\textbf{Time-frequency representation and 2D-CNN}} & \multicolumn{6}{l}{} \\
 & Piczak-CNN (2015)  & \cite{piczak2015cnn} & Mel-spec & 90.20 & 64.50 & 73.70 & --\;\; \\
 & SB-CNN (2017)  & \cite{salamon2017cnn} & Mel-spec & -- & -- & 79.00 & --\;\; \\
 & GoogLeNet (2017)  & \cite{boddapati2017classifying} & Mel-spec, MFCC, CRP & 86.00 & 73.00 & -- & 93.00\,\tnote{2} \\
 & Piczak-CNN (2017)  & \cite{agrawal2017teo} & (TEO-)GT-spec & -- & 81.95 & --\;\; & 88.02\,\tnote{3}\\
 & Piczak-CNN (2017)  & \cite{tak2017pefbe} & (PE)FBE & -- & 84.15 & -- & --\;\; \\
 & VGG-like CNN + mix-up (2018)  & \cite{zhang2018mixup} & Mel-, GT-spec & 91.70 & 83.90 & 83.70 & --\;\; \\
 & VGG-like CNN + Bi-GRU + att. (2019)  & \cite{zhang2019crnn} & GT-spec & 94.20 & 86.50 & -- & --\;\; \\
\addlinespace[0.5em]
 & TSCNN-DS (2019)  & \cite{su2019tscnnds} & Mel-spec, MFCC, CST & -- & -- & -- & 97.20\;\; \\
 & LMCNet (2019)  & \cite{su2019tscnnds} & Mel-spec, CST & -- & -- & -- & 95.20\;\; \\
 & LMCNet (no aug.) & \emph{reproduced}\,\tnote{4} & Mel-spec, CST\,\tnote{5} & -- & -- & \emph{74.04} & \emph{94.00}\;\; \\
\addlinespace[0.5em]
 & TFNet (2019) & \cite{wang2019tfnet} & Mel-spec & 95.80 & 87.70 & -- & 88.50\;\; \\
 & TFNet (no aug.) (2019) & \cite{wang2019tfnet} & Mel-spec & 93.10 & 86.20 & -- & 87.20\;\; \\
 & TFNet (no aug.) & \emph{reproduced}\,\tnote{6} & Mel-spec\,\tnote{7} & -- & \emph{79.45} & \emph{78.50} & \emph{96.69}\;\; \\
\cmidrule(l){1-8} 
\parbox[t]{2mm}{\multirow{6}{*}{\rotatebox[origin=c]{90}{\underline{Ours}}}} & \multicolumn{7}{l}{\textbf{ESResNet}} \\
 & from scratch & & log-power spec & 92.50 & 81.15 & 81.31 & (96.74)\;\; \\
 & ImageNet pre-trained & & log-power spec & 96.75 & 90.80 & 84.90 & (98.18)\;\; \\
 & \multicolumn{7}{l}{\textbf{ESResNet-Attention}} \\
 & from scratch & & log-power spec & 94.25 & 83.15 & 82.76 & (96.83)\;\; \\
 & ImageNet pre-trained & & log-power spec & \textbf{97.00} & \textbf{91.50} & \textbf{85.42} & (\textbf{98.84})\;\; \\
\bottomrule
\end{tabularx}
\begin{tablenotes}
    The table shows a comprehensive overview of the achieved accuracy in percent. Numbers on the ESC and UrbanSound8K (US8K) dataset are as originally reported in the source. If not indicated otherwise, we differentiate into the US8K official or unofficial column according to our findings.
\end{tablenotes}
\begin{tablenotes}
\footnotesize
    \underline{Abbreviations}:\:
    \item FBE: FilterBank Energies \cite{sailor2017convrbm};
    \item spec: spectrogram;
    \item MFCC: Mel-Frequency Cepstral Coefficients \cite{jiang2002spectral};
    \item CRP: Cross Recurrence Plot \cite{marwan2002crp};
    \item TEO: Teager's Energy Operator \cite{kaiser1993teo};
    \item GT: GammaTone \cite{slaney1993gammatone};
    \item (PE)FBE: (Phase-Encoded) FilterBank Energies \cite{tak2017pefbe};
    \item CST: Chromagram, Spectral contrast and Tonnetz \cite{su2019tscnnds}.
\end{tablenotes}
\begin{tablenotes}
\footnotesize
    \underline{Comments}:\:
    \item[1] ``The audio files were segmented into 16,000 samples and successive frames have 50\:\% of overlapping. Ten percent of the dataset was used as validation set and 10\:\% percent of the dataset was also used as test set. Each network was trained with 80\:\% of the dataset''  \cite{abdoli2019cnn1d};
    \item[2] ``We used 5-fold cross validation'' \cite{boddapati2017classifying};
    \item[3] Determined by \cite{zhang2018mixup};
    \item[4] Full re-implementation (based on description in \cite{su2019tscnnds});
    \item[5] Computed according to \cite{su2019tscnnds};
    \item[6] Partial re-implementation (based on temporarily available code (incomplete) from \cite{wang2019tfnet});
    \item[7] Code from \cite{wang2019tfnet} used.
\end{tablenotes}
\end{threeparttable}
\end{table*}

\begin{table}[ht]
\caption{Evaluation Results of the \mbox{ESResNet(-Attention)} Model on Mono and Stereo Input (Accuracy, \%)}
\label{tbl:results_us8k}
\ra{1.2}
\begin{tabularx}{\linewidth}{Xccc}
\toprule
\multicolumn{1}{l}{\multirow{2}{*}{Model}} & \multicolumn{1}{l}{\multirow{2}{*}{ImageNet pre-trained}} & \multicolumn{2}{c}{UrbanSound8K} \\ \cmidrule(l){3-4} 
 & & \multicolumn{1}{c}{mono} & \multicolumn{1}{c}{stereo} \\ \midrule
\multicolumn{1}{l}{\multirow{2}{*}{ESResNet}} & No & 79.91 & 81.31 \\
& Yes & 83.59 & 84.90 \\
\multicolumn{1}{l}{\multirow{2}{*}{ESResNet-Attention}} & No & 81.00 & 82.76 \\
 & Yes & \textbf{84.21} & \textbf{85.42} \\ \bottomrule
\end{tabularx}
\end{table}

As can be observed in \autoref{tbl:results}, our presented approaches out-perform all previous approaches in a fair comparison.

\subsection{ImageNet Weights vs. Random Weights} \label{ssec:imagenet_vs_random}
As we discussed in the \autoref{ssec:datasets}, the amount of available training samples plays a crucial role for deep learning models.
In this work, we compared performance differences between a model that was trained from scratch and one that was pre-trained on the ImageNet dataset and then fine-tuned.
The largest relative change can be observed on the \mbox{ESC-50} dataset (from 81.15\,\% to 90.80\,\%, \mbox{ESResNet}) as it presents a challenging problem in conjunction with a restricted number of training samples.
We still find strong improvements on the \mbox{ESC-10} (from 92.50\,\% to 96.75\,\%, \mbox{ESResNet}) and US8K dataset (from 79.91\,\% to 83.59\,\%, \mbox{ESResNet}).

\subsection{Stereo vs. Mono} \label{ssec:stereo_vs_mono}
Further, despite the availability of stereo recordings in the US8K, we identified, that the competing previous models only consider single-channel audio.
As described, we use a Siamese-like extension to the vanilla input processing of the \mbox{ResNet-50} network in order to enable our \mbox{ESResNet} architecture to process multi-channel inputs where possible (US8K in \autoref{tbl:results}).
Further,
\autoref{tbl:results_us8k} presents a comparison of results of our model achieved on mono and stereo inputs.
The results show that between-channel difference provides useful information that allows to out-perform previous \mbox{state-of-the-art} results on the \mbox{US8K} dataset even without the use of additional attention blocks.
For instance, the \mbox{ESResNet} model trained from scratch is able to achieve accuracy of 79.91\,\% on the \mbox{US8K} dataset using mono audio as an input, however the use of stereo input allows to classify 81.31\,\% of the test samples correctly whereas the extension by attention blocks (\mbox{ESResNet-Attention}) provides a smaller performance gain when operating only on mono input (81.00\,\%).
A similar situation can be observed in the case of our \mbox{ESResNet} model that was pre-trained on the ImageNet dataset \cite{deng2009imagenet}.
The use of stereo input for the \mbox{ESResNet} model out-performs (84.90\,\%) the vanilla model on mono input (83.59\,\%) as well as the attention-boosted model on mono input (84.21\,\%).

\subsection{Attention-boosted vs. Vanilla} \label{ssec:attention_vs_vanilla}
Combining a powerful visual model and descriptive time-frequency representation (ESResNet) already allows us to out-perform previous results.
However, further improvement is possible by including attention (\mbox{ESResNet-Attention}, \autoref{fig:esresnet_a}).
The use of the attention blocks allows us to out-perform previous \mbox{state-of-the-art} results on all three datasets (\mbox{ESC-50 / -10} and \mbox{US8K}) achieving 91.50\,\%, 97.00\,\% and 84.21\,\%, respectively.
Additionally, the combination of stereo input and attention blocks provides further improvement of the achieved accuracy on the \mbox{US8K}, allowing the  \mbox{ESResNet-Attention} model to achieve a new highest \mbox{state-of-the-art} accuracy of 85.42\,\%.

\subsection{Official and Unofficial Splits and Reproducibility Problems} \label{ssec:off_vs_unoff}

As stated in the \autoref{ssec:reimpl}, we reproduced the approaches presented in \cite{su2019tscnnds} and \cite{wang2019tfnet}.

The performance achieved by the re-implemented \mbox{LMCNet} model on the US8K (\autoref{tbl:results}) allows us to attribute the results of \cite{su2019tscnnds} to those that did not perform evaluation on the official split.

For TFNet, we re-ran the temporarily available code for the ESC-50 dataset (without data-augmentation).
We then slightly adapted the code to also run it on the US8K.
In both cases, we surprisingly reached significantly lower accuracies than stated by the authors \cite{wang2019tfnet}.
However, when running their code on a completely random unofficial US8K split, we achieved significantly higher results than previously reported.
We conclude from this, that either, the shared code lacks crucial steps for reproducibility of the reported results, or that the authors neither ran their experiments on the official nor a completely random unofficial US8K split.

In order to roughly quantify the influence of an unofficial (random) splitting strategy on our results, we also report them in \autoref{tbl:results}.
To point out, that these very high numbers do not constitute a basis for fair comparison, we put them in parenthesis.

\section{Conclusion} \label{sec:conc}
In this work we demonstrated how a well-known visual domain model could successfully be applied to Enviromental Sound Classification.
Being applied in conjunction with regular log-power spectrograms, our \mbox{ESResNet} model is able to perform competitive to humans (\mbox{ESC-50}), whereas pre-training on the \mbox{ImageNet} dataset already allows us to out-perform all current \mbox{state-of-the-art} methods.
We also showed that the presence of multiple channels in the input signal gives an additional performance gain on the \mbox{UrbanSound8K} dataset with only minor architectural changes (Siamese-like processing).
Further improvement is possible with the help of attention blocks supporting the network in focusing on the relevant parts of its input in time and frequency domain (\mbox{ESResNet-Attention}).
Such a configuration reached the highest accuracy and out-performed all previous \mbox{state-of-the-art} models significantly in a fair comparison on the ESC-50 and UrbanSound8K datasets.

Finally, we highlighted the importance of the strict adherence to the evaluation procedure, demonstrated the influence of a random splitting strategy on evaluation results on the \mbox{UrbanSound8K} dataset and differentiated previously reported results into official and unofficial splits.
For reproducibility, we provide all code, also including our re-implementations of models that were sadly published without code before.

In the future, we would like to investigate learning time-frequency representations instead of using the current fixed feature extraction.
Also as we have seen that ImageNet helps in the initialization of our model, we would like to investigate which classes benefit more and which less from domain transfer.

\section*{Acknowledgments}
This work was supported by the TU Kaiserslautern CS PhD scholarship program, the BMBF project DeFuseNN (Grant 01IW17002) and the NVIDIA AI Lab (NVAIL) program.
Further, we thank all members of the Deep Learning Competence Center at the DFKI for their comments and support. % comment line out if no space left

\bibliographystyle{IEEEtran}
\bibliography{references}

% that's all folks
\end{document}